\documentclass[11pt,a4paper]{article}

\usepackage[T1]{fontenc}
\usepackage{mathptmx}                  
\usepackage[margin=1in]{geometry}
\usepackage{amsmath,amssymb}
\usepackage{graphicx}
\usepackage{booktabs}
\usepackage{caption}
\usepackage{url}
\usepackage[hidelinks]{hyperref}

\title{\textbf{An Uncertainty-Aware Hybrid Mathematical--Machine-Learning Model for Smart Irrigation Decision Support}}
\author{Andrea Scariolo\\ \textit{Independent Researcher}\\ \texttt{andrea.scariolo@icloud.com}}
\date{}

\begin{document}
\maketitle

\begin{figure}[htbp]
\centering
\includegraphics[width=1.00\linewidth]{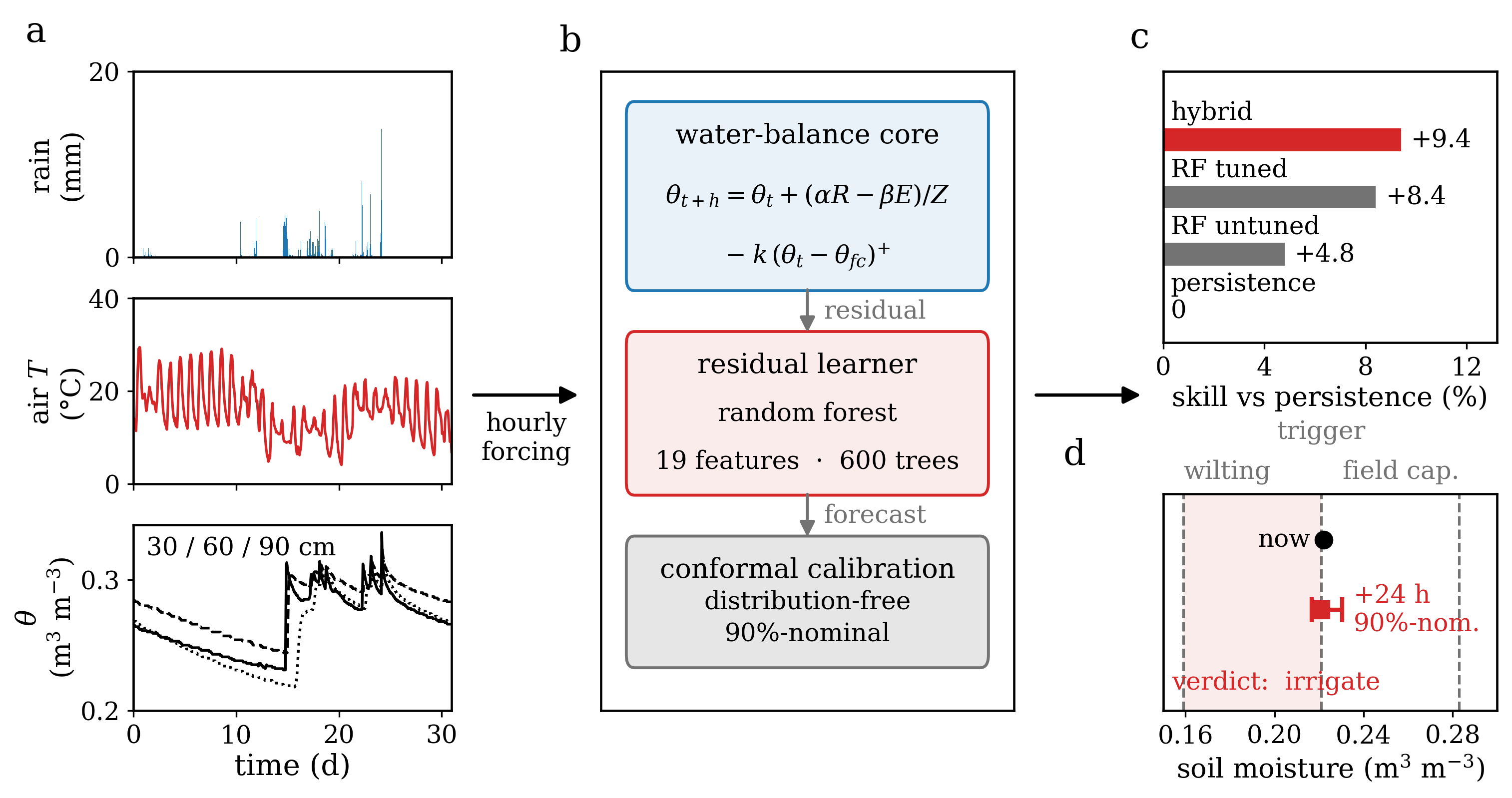}
\caption{(a) The measured hourly streams the model consumes at the Torano station: rainfall, air temperature and volumetric soil moisture at 30, 60 and 90\,cm, shown here over one month of the record. (b) The two-stage architecture. A transparent four-parameter water-balance core produces a physically grounded forecast, a tuned Random Forest corrects its residual, and conformal calibration attaches a distribution-free 90\,\%-nominal interval. (c) Skill relative to persistence at the 24\,h horizon for the models actually run; the hybrid leads at +9.4\,\%. (d) The forecast and its interval turned into a decision for one real issue time in the test partition, 6 April 2010 at 16:00. Measured moisture sits just above the management trigger, and the 90\,\%-nominal interval of the 24\,h forecast already reaches below it, so the risk-aware rule calls for irrigation while the point forecast alone would not.}
\label{fig:1}
\end{figure}

\begin{abstract}
Agriculture accounts for roughly 70\,\% of global freshwater withdrawals, yet irrigation is still commonly scheduled reactively, without a forecast of where soil moisture is heading or any statement of confidence in that forecast. Data-driven soil-moisture models are accurate but opaque and point-valued, whereas physically based water-balance models are transparent but carry large structural error; neither alone supports a defensible irrigation decision under uncertainty. This study coupled the two and carried uncertainty through to the decision. A four-parameter water-balance core, calibrated on training data only, was corrected by a Random Forest that learned nothing but the physical residual; distribution-free conformal prediction attached 90\,\%-nominal prediction intervals; and a risk-aware rule converted the interval lower bound into an irrigation trigger. The framework was evaluated on three years of hourly in-situ measurements from a rainfed Mediterranean cropland station under a strict chronological split, scored against persistence, and swept across lead times from one hour to one week. At the 24\,h horizon the tuned hybrid reached RMSE 0.00925\,$\mathrm{m^{3}\,m^{-3}}$, a skill of +9.4\,\% over persistence and roughly double the best of nine benchmarked baselines, of which only the Random Forest beat persistence at all. Skill was strongly horizon-dependent and, contrary to expectation, did not grow with lead time: it peaked at +27.4\,\% at three hours and decayed to +1.2\,\% at one week, a decay consistent with unknown future rainfall coming to dominate the error budget for this dataset and feature set. Conformalised quantile regression was better calibrated and 11\,\% sharper than constant-width conformal prediction (Winkler score 0.0405 versus 0.0517), widening in wet conditions and narrowing in dry ones, and its mean width grew 40-fold across the horizon range. In the decision-relevant regime the risk-aware rule raised the share of management-threshold crossings detected in advance from 0.905 to 1.000, at a precision cost of 0.975 to 0.950 and a 3.7\,\% increase in a notional refill-to-field-capacity water proxy, so the rule buys earlier detection with water and selectivity rather than saving water. Beyond 72\,h the point forecast fell below the no-forecast rule while the interval-based rule did not. Uncertainty quantification therefore governs the lead time over which forecast-driven irrigation advice remains trustworthy, and in this configuration transparency in the physical layer cost no measurable accuracy.
\end{abstract}

\noindent\textbf{Keywords:} soil moisture forecasting; hybrid physics--machine learning; conformal prediction; irrigation decision support; interpretability; Mediterranean agriculture

\section{Introduction}

Agriculture accounts for approximately 70\,\% of global freshwater withdrawals, and irrigation is its largest single component (Food and Agriculture Organization of the United Nations [FAO], n.d.). Pressure on that water is rising fastest in semi-arid regions, where Mediterranean cropping systems combine hot dry summers with highly variable rainfall. Machine learning has become central to agricultural decision-making across crop, water and soil management (Liakos et al., 2018), and sensor-driven soil-moisture prediction within Internet-of-Things irrigation systems is by now an established task (Goap et al., 2018). Despite this, on-farm irrigation is still typically scheduled reactively, once the soil is already dry, or on fixed calendars. Two capabilities are absent from most operational decision support: a prediction that is simultaneously accurate and \emph{auditable}, and an \emph{uncertainty estimate} on which a grower can act. A recommendation that cannot be traced to physically meaningful quantities is difficult to trust, and a single predicted number carries no information about when it should be distrusted.

Soil-moisture prediction has advanced rapidly, but along four largely separate lines. \textbf{Deep data-driven forecasting} now dominates the literature. Wang et al. (2024) benchmarked ten deep architectures and used Shapley analysis to explain how each exploited the input signal, finding attention mechanisms and adversarial training most effective. Zheng et al. (2024) proposed a GRU--Transformer hybrid whose advantage over conventional learners was largest at one- to two-day horizons, and Ahmed et al. (2021) combined empirical mode decomposition with a convolutional--recurrent network for multi-step forecasts. Xu et al. (2023) coupled deep learning to sub-seasonal process-model forecasts and improved skill at both surface and root-zone depths. A parallel remote-sensing branch drives prediction from satellite and reanalysis products (Bakhshian et al., 2025; Koohikeradeh et al., 2025; Nguyen et al., 2022).

\textbf{Hybrid physics--machine-learning} models are the closest relatives of the present work. Li et al. (2024) injected the output of physically based land-surface models into deep networks and showed the hybrid outperformed pure deep learning, particularly at long lead times and for drought events. Izquierdo-Sanz and Molt\'o (2026) inverted a physical scattering model from Sentinel imagery and corrected it with machine learning in irrigated Mediterranean citrus orchards, the study closest to this one in setting. El Rhadiouini et al. (2026) proposed a physics-guided baseline with a learned correction for retrieval in complex terrain.

\textbf{Explainable artificial intelligence} has been applied systematically to this task. Huang et al. (2023) built the most complete treatment, combining permutation importance and accumulated local effects for global interpretation with Shapley values, local surrogate models and individual conditional expectation for local interpretation, and recovered physically meaningful drivers of soil-moisture drought. Mallik et al. (2025) and Mohan et al. (2025) applied comparable tooling to sensor-fusion and yield-prediction problems.

Finally, a small \textbf{uncertainty-aware} literature exists. Zheng et al. (2026) quantified predictive uncertainty by Monte-Carlo dropout over a deep spatio-temporal model and folded moisture deviation, spatial variance and interval width into a genetic-algorithm-optimised irrigation index, reporting water savings against fixed-threshold scheduling. Tandon et al. (2022) attached particle-filter uncertainty to a neural forecast out to five days. Recent reviews (Ajith et al., 2025; Islam et al., 2025; Khaliq et al., 2025; Settu \& Ramaiah, 2025; Taheri et al., 2025) independently identify transparency, uncertainty quantification and spatial transferability as the field's open problems.

\subsection{Gaps in the current literature}

Read together, this body of work leaves five specific gaps.

\emph{G1. Accuracy without auditability.} The most accurate forecasters are black boxes emitting a single number. An irrigation recommendation must be traceable to physically meaningful quantities, and a point estimate cannot signal when it should not be believed.

\emph{G2. Hybrid models stop at estimation.} Existing physics--machine-learning couplings target retrieval or estimation of the current state rather than a forecast tied to a decision, and none pairs the physical core with calibrated predictive intervals or an explicit irrigation rule. Their physical component is typically a full land-surface or scattering model, which restores the opacity the hybrid was meant to remove.

\emph{G3. Explainability serves the predictor, not the decision.} Explainable-AI work in this field interprets why a model predicted a value and stops there; the identified drivers are not carried through to justifying why an irrigation action was recommended.

\emph{G4. Uncertainty is model-based and rarely reaches the decision.} Monte-Carlo dropout and particle filtering depend on the model being correctly specified and carry no finite-sample coverage guarantee. Distribution-free conformal prediction is essentially absent from operational soil-moisture forecasting, and \emph{conditional} coverage, whether an interval is trustworthy in wet and dry conditions separately, is not audited.

\emph{G5. Evaluation is optimistic and horizon-bound.} Persistence, an extremely strong baseline for an autocorrelated variable at short lead, is frequently omitted, so reported accuracy does not establish that a model adds information. Most studies fix a single lead time, leaving the range over which a forecast is actually useful undetermined, and negative results are seldom reported.

\subsection{Contribution and research questions}

This study addressed the five gaps within a single auditable pipeline. A transparent water-balance core with four interpretable parameters, calibrated on training data alone, provided the physically grounded forecast (G1, G2). A machine-learning model corrected only its residual, permutation and Shapley analysis identified the drivers of the predictive component, and the irrigation action itself was left interpretable by construction through an explicit threshold rule rather than inferred from feature attribution (G3). A distribution-free conformal layer attached 90\,\%-nominal prediction intervals whose conditional coverage was audited by regime (G4). A risk-aware rule converted the interval lower bound into an irrigation trigger. The framework was validated on real Mediterranean in-situ data under a strict chronological split, scored throughout against persistence, evaluated across lead times from one hour to one week, and reported together with its negative results (G5).

Three research questions were addressed:

\begin{itemize}
\item \textbf{RQ1.} Does the hybrid mathematical--machine-learning model predict soil moisture, and the derived   irrigation need, better than standard baselines under time-aware evaluation?
\item \textbf{RQ2.} Can the model produce interpretable decisions and identify the principal driving variables?
\item \textbf{RQ3.} Does calibrated predictive uncertainty improve the early detection of management-threshold   crossings, and what additional cost in notional water and in selectivity does that improvement   carry?
\end{itemize}

\section{Materials and Methods}

\subsection{Study site and data}

Data were obtained from the CALABRIA network of the International Soil Moisture Network, in Calabria, southern Italy, a Mediterranean climate under the K\"oppen Csa classification. Analysis focused on the Torano station, whose land cover is recorded as rainfed cropland, which makes an irrigation trigger agronomically meaningful. The station provided hourly volumetric soil moisture at 0.3, 0.6 and 0.9\,m depth, air temperature at 2\,m, and rain-gauge precipitation. Static soil properties from the Harmonized World Soil Database gave a saturation of 0.44\,$\mathrm{m^{3}\,m^{-3}}$ and a clay-loam texture (22\,\% clay, 42\,\% sand, 36\,\% silt) in the upper 0.3\,m.

An exploratory analysis conducted before modelling returned a pre-registered positive verdict against four criteria: coverage, physical plausibility, a detectable rainfall response, and a structured water-balance residual. After cleaning, the 30\,cm series was approximately 99\,\% valid, all values fell within the physically admissible range, the correlation between seven-day accumulated rainfall and 30\,cm moisture was 0.27, and the residual of a simplified water balance was significantly non-white by the Ljung--Box test. The last of these was the empirical justification for adding a learned corrector rather than refining the physical model alone.

\subsection{Preprocessing}

Native station files were parsed onto a regular hourly grid. The $-$9999 missing-value sentinel and all records whose International Soil Moisture Network quality flag differed from "good" were set to missing. Remaining gaps were forward-filled for at most two hours; longer gaps were left missing and the affected rows were subsequently dropped. The window in which soil moisture, temperature and rainfall co-existed ran from 6 June 2007 to 31 May 2010 and defined the study period.

\subsection{Feature engineering and evaluation protocol}

The prediction target was volumetric soil moisture at 30\,cm, \emph{h} hours ahead. Nineteen predictors were constructed, all of them past-only quantities available at issue time: soil moisture at the current hour and at lags of 1, 3, 6, 12 and 24\,h; moisture at the two other depths; air temperature and its 24\,h mean; rolling rainfall sums over 6, 24, 72 and 168\,h; the 24\,h change in moisture; and sine and cosine harmonics of day-of-year and hour-of-day. The same nineteen predictors were retained at every horizon so that comparisons across lead time were not confounded by a changing feature set. After construction, 24\,752 complete samples remained at the 24\,h horizon.

Evaluation followed a strict chronological 70/15/15 split: 17\,326 training samples ending 13 July 2009, 3\,713 validation samples, and 3\,713 test samples spanning 16 December 2009 to 30 May 2010. No shuffling was applied. The test partition was not accessed during training, hyperparameter search, or conformal calibration, so no information from it could leak into any fitted quantity.

\subsection{Physical core}

A single-layer hourly water balance was adopted for the 30\,cm horizon, with an effective storage depth \emph{Z} = 300\,mm:

\begin{equation}
SM(t+h) = SM(t) + \frac{\alpha R_h - \beta ET_h}{Z} - k \max\left(SM(t) - \theta_{fc},\ 0\right)
\end{equation}

Here \emph{R\_h} is rainfall accumulated over the forecast window, \emph{ET\_h} is Hargreaves reference evapotranspiration accumulated over the same window, computed from temperature alone with extraterrestrial radiation evaluated at the station latitude of 39.5$^\circ$ N following FAO-56, and the final term is a linear drainage and runoff leakage active above a field-capacity threshold $\theta_{\mathrm{fc}}$. The four parameters ($\alpha$, $\beta$, \emph{k}, $\theta_{\mathrm{fc}}$) were calibrated by exhaustive grid search minimising training-set RMSE; no test or validation data entered the calibration.

Two forcing modes were distinguished. In the \textbf{deployable} mode, future rainfall is unknown at issue time and $\alpha$ was therefore fixed at zero. In a \textbf{perfect-forcing ablation}, observed future rainfall was supplied to the core in order to bound the improvement attainable from a perfect rainfall forecast. The ablation is not a deployable configuration and is reported separately throughout.

\subsection{Machine-learning residual corrector}

The machine-learning layer did not predict soil moisture. It predicted the residual \emph{r(t)} = \emph{SM(t+h)} $-$ $SM_{\mathrm{phys}}(t+h)$, that is, precisely the component the bucket model fails to capture: infiltration delays, soil heterogeneity, and unmodelled losses. The final forecast was the sum of the physical estimate and the learned correction, which keeps the physical terms readable individually and confines the opaque component to a correction of known magnitude.

A Random Forest was used as the corrector. Its hyperparameters were selected by Bayesian optimisation (Shahriari et al., 2016) with a tree-structured Parzen estimator over 30 trials, minimising validation RMSE; the selected configuration used 600 trees, maximum depth 17, a maximum-feature fraction of 0.20, a minimum of seven samples per leaf and eight per split. The same configuration was reused for the direct-prediction baseline and for the residual corrector, and at every horizon, which is stated as a limitation in Section 4.5.

\subsection{Uncertainty layer}

Prediction intervals at the 90\,\% nominal level were calibrated on the validation partition, which is strictly earlier in time than the test partition. Neither method is assumed to reach that level on the test partition: the empirical coverage each attains there is itself a measured quantity, reported in Section 3.5, and both are therefore described throughout as 90\,\%-\emph{nominal} rather than as calibrated.

Two methods were compared, and they attach to different points of the pipeline. \textbf{Split conformal prediction} was applied to the hybrid point forecast: the conformity score was the absolute residual of the tuned hybrid on validation, its $\lceil (n+1)(1-\alpha) \rceil / n$ empirical quantile \emph{q} was taken, and the interval was $\hat{y}(t)$ $\pm$ \emph{q}, symmetric and of constant width. \textbf{Conformalised quantile regression} (CQR) was applied to the \emph{physical residual}: two gradient-boosted quantile regressors were fitted on the training partition to \emph{r} = \emph{SM}(\emph{t + h}) $-$ $SM_{\mathrm{phys}}(t+h)$ at the $\alpha/2$ and $1-\alpha/2$ quantiles, giving $\hat{q}_{\mathrm{lo}}(x)$ and $\hat{q}_{\mathrm{hi}}(x)$; the conformity score $E = \max(\hat{q}_{\mathrm{lo}} - r,\ r - \hat{q}_{\mathrm{hi}})$ was evaluated on validation and its corresponding quantile \emph{Q} taken; and the interval was

\begin{equation}
\bigl[\,SM_{\mathrm{phys}}(t+h) + \hat{q}_{\mathrm{lo}}(x) - Q,\ \ SM_{\mathrm{phys}}(t+h) + \hat{q}_{\mathrm{hi}}(x) + Q\,\bigr]
\end{equation}

that is, the physical forecast plus a conformally widened, input-dependent band on its residual. This construction is what makes the CQR width state-dependent: the quantile regressors see the same rainfall and antecedent-wetness features as the corrector, so the band responds to the conditions under which the physical core is known to fail.

Interval quality was assessed by marginal coverage, mean width as a measure of sharpness, and the Winkler interval score, which penalises both excessive width and missed observations. The principal diagnostic was \emph{conditional} coverage: coverage and width were recomputed separately for wet and dry regimes, defined by 24\,h accumulated rainfall above or below its median at issue time, and by season.

\subsection{Irrigation decision rule}

Thresholds were derived from the data rather than assumed. Field capacity was taken as the calibrated physical threshold, $\theta_{\mathrm{fc}}$ = 0.283\,$\mathrm{m^{3}\,m^{-3}}$. A wilting proxy $\theta_{\mathrm{wp}}$ = 0.159\,$\mathrm{m^{3}\,m^{-3}}$ was taken as the fifth percentile of observed moisture and cross-checked against a clay-loam pedotransfer estimate of approximately 0.12\,$\mathrm{m^{3}\,m^{-3}}$. The management trigger followed the standard 50\,\% management-allowable depletion rule, $\theta_{\mathrm{trig}}$ = $\theta_{\mathrm{fc}}$ $-$ 0.5($\theta_{\mathrm{fc}}$ $-$ $\theta_{\mathrm{wp}}$) = 0.221\,$\mathrm{m^{3}\,m^{-3}}$. The adverse event was defined as a \textbf{management-threshold crossing}, that is, moisture at \emph{t + h} falling below $\theta_{\mathrm{trig}}$. This is an empirically defined soil-moisture condition derived from the data, not an observed physiological crop-stress event; the dataset contains no plant measurement, so the event should be read as a potential moisture-stress condition that a manager would act on, and no claim is made that crop stress in fact occurred.

Three rules were compared. The \textbf{reactive} rule, representing current practice without a forecast, irrigated when \emph{current} moisture fell below the trigger. The \textbf{forecast} rule irrigated when the \emph{predicted} moisture fell below it. The \textbf{risk-aware} rule, which constitutes the decision support system, irrigated when the \emph{lower bound of the 90\,\%-nominal interval} fell below it. This is a conservative decision rule at the chosen nominal coverage level: it acts whenever the trigger is not excluded by the interval, and is not a statement that the probability of crossing the trigger exceeds a particular value, since the attained coverage is itself a measured quantity (Section 3.5).

Evaluation was open-loop. Because the dataset contains no irrigation actuation, post-irrigation soil state could not be simulated; the analysis therefore measured decision quality and a notional water proxy rather than a closed water balance. The proxy is defined explicitly as

\begin{equation}
W = \sum_{t \in A} \max\bigl(\theta_{fc} - SM(t),\ 0\bigr)\, Z
\end{equation}

where \emph{A} is the set of test hours at which the rule raises an alarm and \emph{Z} = 300\,mm is the effective storage depth: the depth of water that would be required to refill the profile from its observed state at issue time to field capacity. Every alarm hour of the 3\,713-hour test partition is counted independently, without actuation and without updating the soil state, so the aggregate values in Table 6 are an accounting device for comparing rules on a common test set and are \textbf{not} field irrigation depths; only the ratios between rules are meaningful. This framing is a limitation of the data, not a modelling choice, and is discussed in Section 4.5.

Because steady wet and steady dry states are trivially handled by any rule, a \textbf{decision-relevant regime} was defined for diagnostic purposes as those hours in which moisture was declining and within 0.02\,$\mathrm{m^{3}\,m^{-3}}$ of the trigger. This isolates the cases in which irrigation timing is genuinely difficult.

\subsection{Baselines and metrics}

Nine models were benchmarked under the identical chronological split: persistence, ridge regression, a decision tree, a Random Forest, gradient boosting (XGBoost), Gaussian-process regression, a multilayer perceptron, and long short-term memory and gated recurrent unit sequence models. All nine are reported in Table 1. Regression performance was reported as RMSE, mean absolute error and the coefficient of determination. The headline metric throughout was \textbf{skill relative to persistence}. Let \emph{T} denote the set of issue times contained in the test set and \emph{N} = |\emph{T}| its cardinality. For an issue time $t \in T$, write \emph{y}(\emph{t}) = $\theta$(\emph{t} + \emph{h}) for the value observed at the forecast horizon, $\hat{y}(t)$ for the forecast under evaluation, and \emph{p}(\emph{t}) = $\theta$(\emph{t}) for the persistence forecast, that is, the last value observed at issue time propagated unchanged. With the root mean squared error of a forecast \emph{f} taken over the same set,

\begin{equation}
\mathrm{RMSE}(f) = \sqrt{\frac{1}{N}\sum_{t \in T}\bigl(f(t) - y(t)\bigr)^2}
\end{equation}

skill at horizon \emph{h} is defined as

\begin{equation}
S(h) = 1 - \frac{\mathrm{RMSE}(\hat{y})}{\mathrm{RMSE}(p)}
\end{equation}

and is reported as a percentage. A value of one denotes a perfect forecast, zero a forecast indistinguishable from persistence, and a negative value a model that removes no error and therefore adds no information; the score is bounded above by one and unbounded below. Because the denominator is itself a function of \emph{h}, skill compares models across horizons only as a fraction of the persistence error attainable at that horizon, never as an absolute error.

Two conventions for this ratio are in use. The classical form is built from mean squared errors rather than their roots, and returns 1 $-$ (1 $-$ \emph{S}(\emph{h}))$^{2}$ for the same pair of forecasts. The root-mean-square form adopted here is the more conservative of the two, and the values reported below are consequently near half of what the squared-error convention would give. Decision performance was reported as precision, recall, F1 score, the number of missed management-threshold crossings and the notional water proxy. Figure 2 summarises the complete pipeline.

\begin{figure}[htbp]
\centering
\includegraphics[width=0.82\linewidth]{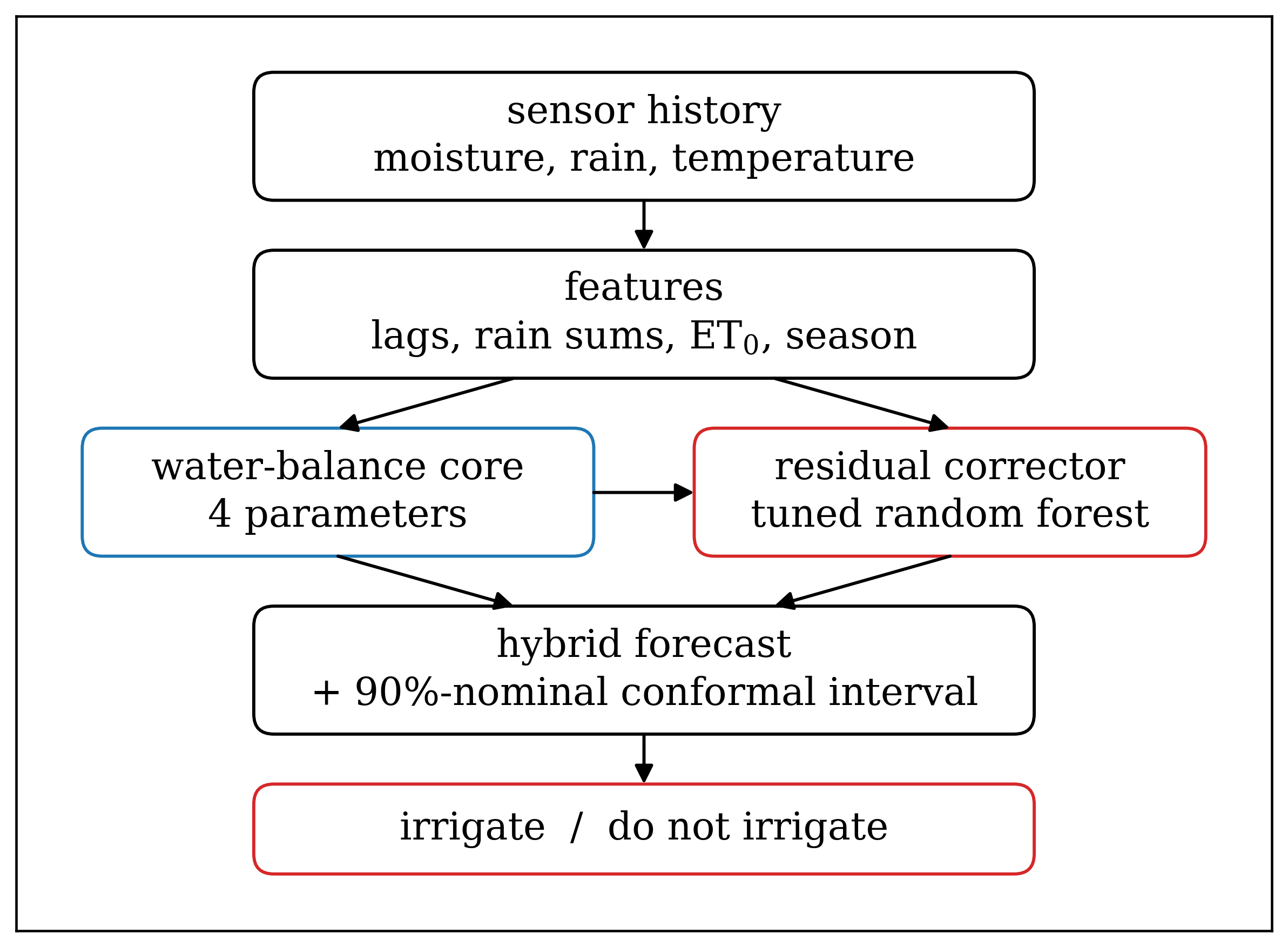}
\caption{Processing chain. Cleaned station records were transformed into past-only features. A transparent water-balance core produced a physically grounded forecast whose residual was corrected by a tuned Random Forest. The resulting hybrid forecast received a calibrated conformal interval, which drove the risk-aware irrigation decision.}
\label{fig:2}
\end{figure}

\section{Results}

\subsection{Baseline comparison}

Under time-aware evaluation the Random Forest was the only baseline that exceeded persistence; every sequence, kernel and linear model performed worse than assuming no change, and among the tree-based learners only the bagged ensemble cleared the bar, gradient boosting falling to $-$5.6\,\% and the single decision tree to $-$11.7\,\% (Table 1). Persistence itself achieved RMSE 0.01021\,$\mathrm{m^{3}\,m^{-3}}$ at the 24\,h horizon. Gaussian-process regression failed severely, with skill of $-$226.8\,\%, and the multilayer perceptron also fell well below the baseline.

\begin{table}[htbp]
\centering
\caption{Nine-model comparison at the 24\,h horizon on the Torano test partition}
\label{tab:1}
\small
\begin{tabular}{lrr}
\toprule
Model & RMSE ($\mathrm{m^{3}\,m^{-3}}$) & Skill vs persistence (\%) \\
\midrule
Random Forest & 0.00972 & +4.8 \\
Persistence & 0.01021 & 0 \\
Ridge regression & 0.01046 & $-$2.4 \\
Gated recurrent unit & 0.01071 & $-$4.9 \\
Long short-term memory & 0.01073 & $-$5.1 \\
Gradient boosting (XGBoost) & 0.01078 & $-$5.6 \\
Decision tree & 0.01141 & $-$11.7 \\
Multilayer perceptron & 0.01387 & $-$35.8 \\
Gaussian-process regression & 0.03336 & $-$226.8 \\
\bottomrule
\end{tabular}
\par\vspace{2pt}
\begin{minipage}{\linewidth}\footnotesize \emph{Note.} Skill is defined by Equation 5. Models are ordered by RMSE.\end{minipage}
\end{table}

\subsection{Hybrid model and boosting}

Constructing the transparent core, correcting its residual, and tuning the corrector approximately doubled the skill of the best baseline (Table 2). The deployable physical core alone reached +3.1\,\% skill. Adding an untuned residual corrector produced +4.6\,\%, matching the tuned-free Random Forest baseline while remaining interpretable term by term. Bayesian optimisation of the corrector raised the direct-prediction Random Forest to +8.4\,\%, the single largest improvement obtained, and the tuned hybrid reached RMSE 0.00925\,$\mathrm{m^{3}\,m^{-3}}$, $R^{2}$ = 0.946 and \textbf{+9.4\,\% skill}, the best deployable configuration.

The calibrated parameters of the deployable core were $\alpha$ = 0 by construction, \textbf{$\beta$ = 0}, \emph{k} = 0.4 and $\theta_{\mathrm{fc}}$ = 0.283\,$\mathrm{m^{3}\,m^{-3}}$. The evapotranspiration weight therefore calibrated to zero.

Under perfect rainfall forcing the physical core reached +23.8\,\% and the hybrid +33.4\,\% with $R^{2}$ = 0.971. These configurations are ablations, not forecasts.

\begin{table}[htbp]
\centering
\caption{Hybrid and boosting results at the 24\,h horizon}
\label{tab:2}
\small
\begin{tabular}{lrrr}
\toprule
Model & RMSE ($\mathrm{m^{3}\,m^{-3}}$) & $R^{2}$ & Skill (\%) \\
\midrule
Hybrid, perfect forcing (ablation) & 0.00680 & 0.971 & +33.4 \\
Physical core, perfect forcing (ablation) & 0.00778 & 0.962 & +23.8 \\
\textbf{Hybrid, tuned corrector} & \textbf{0.00925} & \textbf{0.946} & \textbf{+9.4} \\
Random Forest, tuned & 0.00935 & 0.945 & +8.4 \\
Tri-training, three learners plus unlabelled data & 0.00957 & 0.942 & +6.3 \\
Random Forest, untuned & 0.00972 & 0.940 & +4.8 \\
Hybrid, untuned corrector & 0.00974 & 0.940 & +4.6 \\
Physical core, deployable & 0.00989 & 0.938 & +3.1 \\
Persistence & 0.01021 & 0.934 & 0 \\
\bottomrule
\end{tabular}
\par\vspace{2pt}
\begin{minipage}{\linewidth}\footnotesize \emph{Note.} The physical core was calibrated by grid search restricted to a drainage coefficient $k \le 0.4$; this is the configuration on which the uncertainty and decision layers of Sections 3.5 and 3.6 are built. Section 3.3 reports the effect of widening that range. The tri-training entry is a secondary experiment reported in Section 3.7.\end{minipage}
\end{table}

The margin over a well-tuned black box is narrow: the tuned hybrid at +9.4\,\% and the tuned direct-prediction Random Forest at +8.4\,\% differ by 0.00010\,$\mathrm{m^{3}\,m^{-3}}$ in RMSE. The relevant observation is therefore not that the transparent decomposition wins by a wide margin, but that comparable accuracy is available inside a decomposition whose terms retain a physical reading and to which a 90\,\%-nominal interval and an explicit decision rule can be attached.

\subsection{Forecast horizon}

Skill did not increase monotonically with lead time (Table 3, Figure 3). Absolute error grew monotonically for both persistence and the hybrid (persistence RMSE rose thirteen-fold from one hour to one week), but relative skill traced an inverted U, rising from +13.6\,\% at one hour to a maximum of \textbf{+27.4\,\% at three hours} and decaying to +1.2\,\% at one week.

The sweep recalibrated the physical core at every horizon, and did so over a wider drainage range ($k \le 0.8$) than the main protocol of Section 3.2 ($k \le 0.4$). Under that wider range the core evaluated alone remained within $\pm$1.6\,\% of persistence at every horizon, and at 24\,h it selected $k = 0.60$ and returned RMSE 0.01017\,$\mathrm{m^{3}\,m^{-3}}$, a standalone skill of +0.35\,\% rather than the +3.1\,\% obtained under the restricted grid. The hybrid was unaffected by the change, returning RMSE 0.00925\,$\mathrm{m^{3}\,m^{-3}}$ in both cases, because the residual learner absorbed the difference; the sensitivity of the standalone core to the calibration range is discussed in Section 4.2.

\begin{table}[htbp]
\centering
\caption{Forecast performance versus lead time}
\label{tab:3}
\small
\begin{tabular}{lrrrr}
\toprule
Lead time (h) & RMSE persistence & RMSE hybrid & $R^{2}$ & Skill (\%) \\
\midrule
1 & 0.00178 & 0.00154 & 0.998 & +13.6 \\
\textbf{3} & \textbf{0.00356} & \textbf{0.00258} & \textbf{0.996} & \textbf{+27.4} \\
6 & 0.00524 & 0.00398 & 0.990 & +24.0 \\
12 & 0.00746 & 0.00624 & 0.975 & +16.4 \\
24 & 0.01021 & 0.00925 & 0.946 & +9.4 \\
48 & 0.01347 & 0.01286 & 0.896 & +4.5 \\
72 & 0.01600 & 0.01530 & 0.852 & +4.4 \\
168 & 0.02384 & 0.02355 & 0.647 & +1.2 \\
\bottomrule
\end{tabular}
\end{table}

The physical parameters recalibrated at each horizon over the wider grid behaved systematically. The drainage coefficient \emph{k} increased with lead time from 0.05 at one hour to 0.60 at 24\,h and 0.50--0.65 at 48--168\,h, and the field-capacity threshold drifted downward from 0.283 to 0.263\,$\mathrm{m^{3}\,m^{-3}}$ at the longest horizon. The evapotranspiration weight $\beta$ calibrated to zero \textbf{at every horizon tested, including one week}.

\begin{figure}[htbp]
\centering
\includegraphics[width=1.00\linewidth]{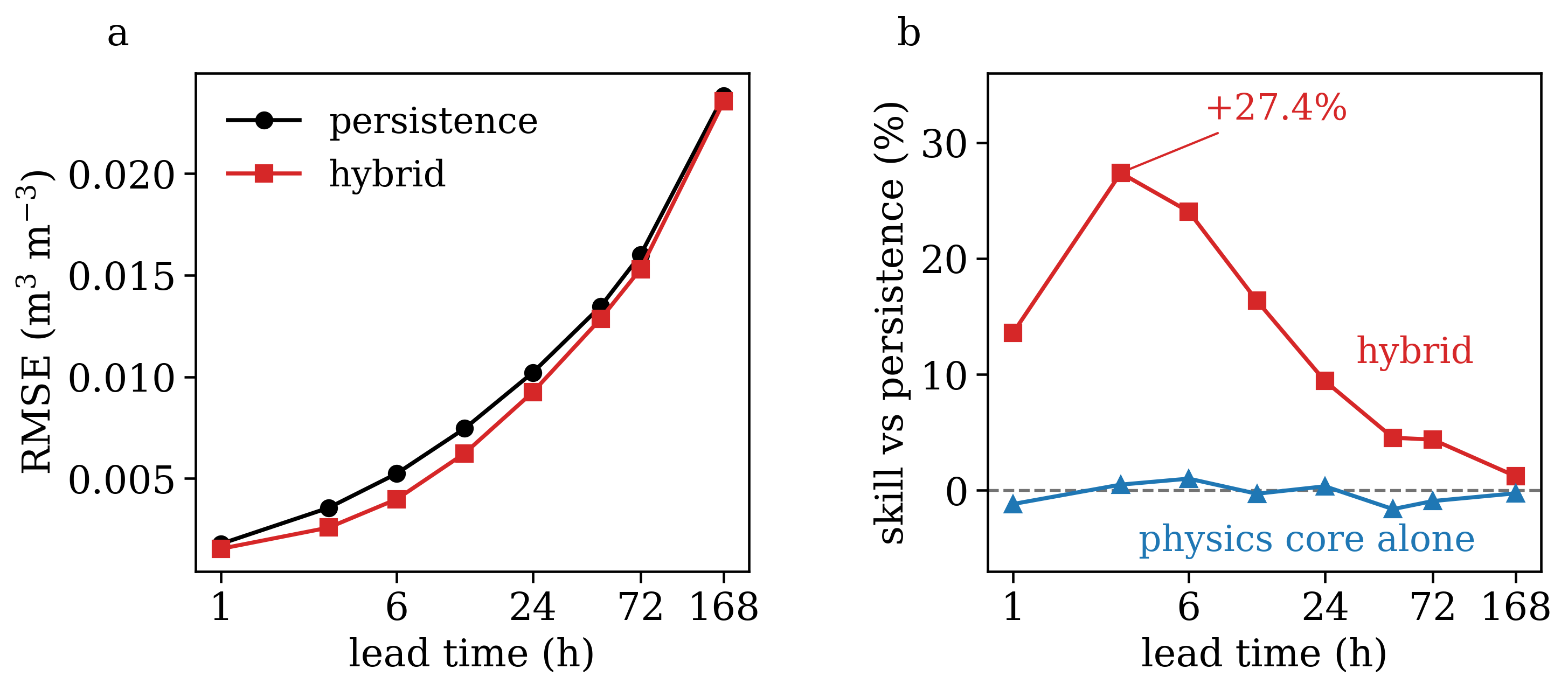}
\caption{Error and skill against lead time. (a) Root-mean-square error of persistence and of the tuned hybrid; both grow monotonically. (b) Skill relative to persistence; the hybrid peaks at three hours and decays towards zero within a week, while the physical core alone remains at the persistence level throughout.}
\label{fig:3}
\end{figure}

\subsection{Driving variables}

Both attribution analyses were computed on the \textbf{tuned direct-prediction Random Forest}, that is, the model whose target is soil moisture at \emph{t} + 24\,h rather than the physical residual. It carries the same nineteen predictors and the same hyperparameter configuration as the residual corrector, so its attributions describe the drivers of the \emph{predictive component} of the pipeline; they are not an explanation of the irrigation action, which is interpretable by construction through the explicit threshold rule of Section 2.7. Shapley values were obtained with an exact tree explainer on a random subsample of 1\,500 test rows.

Permutation importance computed on the held-out partition and Shapley attributions computed on that model agreed on the overall structure (Figure 4). The current state dominated by a wide margin, with a mean absolute Shapley value of 0.0094\,$\mathrm{m^{3}\,m^{-3}}$, followed by the short lags in descending order. The leading non-soil driver was antecedent weekly wetness at 0.0023\,$\mathrm{m^{3}\,m^{-3}}$, followed by the 24\,h mean temperature at 0.0017\,$\mathrm{m^{3}\,m^{-3}}$ and the daily rainfall total at 0.0015\,$\mathrm{m^{3}\,m^{-3}}$. The two methods differed only in their ranking of the rainfall features: permutation importance placed recent six-hour rainfall highest among them, whereas Shapley analysis placed the weekly total highest.

\begin{figure}[htbp]
\centering
\includegraphics[width=1.00\linewidth]{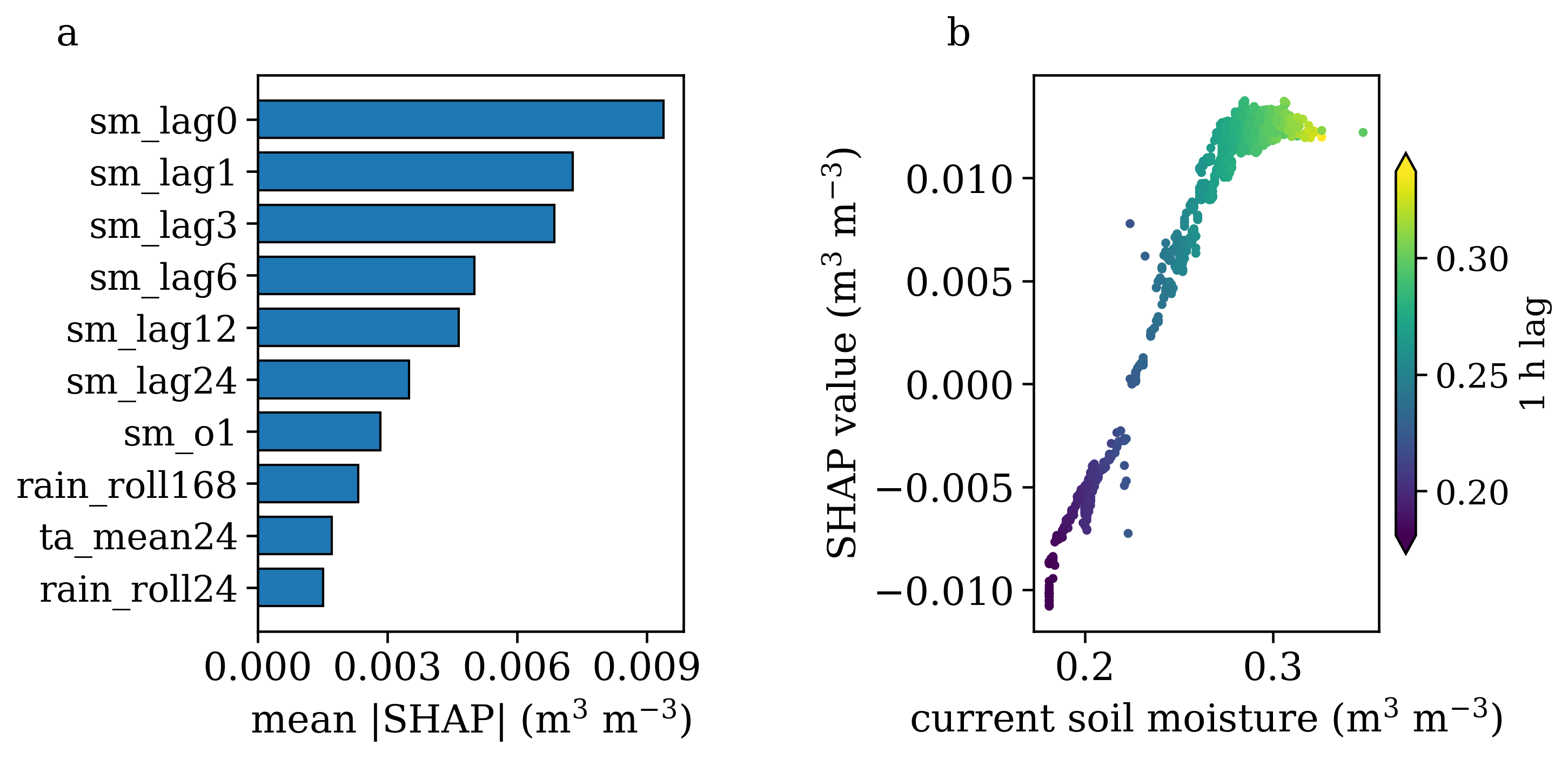}
\caption{Attribution of the tuned direct-prediction Random Forest. (a) Global mean absolute Shapley values for the ten leading predictors. (b) Shapley dependence for the dominant predictor, coloured by the one-hour lag, showing a near-monotone relationship between current moisture and its contribution to the forecast.}
\label{fig:4}
\end{figure}

\subsection{Interval calibration and sharpness}

Both intervals are 90\,\%-nominal by construction, the split conformal band being symmetric about the hybrid point forecast and the CQR band being an input-dependent band on the physical residual added to the physical forecast (Section 2.6). At the 24\,h horizon neither attained the nominal level on the test partition (Table 4). Split conformal prediction achieved 0.826 empirical coverage at a constant width of 0.01939\,$\mathrm{m^{3}\,m^{-3}}$, and CQR achieved 0.875 coverage at a mean width of 0.01725\,$\mathrm{m^{3}\,m^{-3}}$, a 22\,\% lower Winkler score.

\begin{table}[htbp]
\centering
\caption{Prediction-interval quality at the 24\,h horizon, 90\,\% nominal level}
\label{tab:4}
\small
\begin{tabular}{lrrr}
\toprule
Method & Coverage & Mean width ($\mathrm{m^{3}\,m^{-3}}$) & Winkler score \\
\midrule
Split conformal & 0.826 & 0.01939 & 0.05172 \\
\textbf{Conformalised quantile regression} & \textbf{0.875} & \textbf{0.01725} & \textbf{0.04050} \\
\bottomrule
\end{tabular}
\end{table}

The conditional audit separated the two methods more sharply (Table 5). Split conformal prediction over-covered the dry regime at 0.906 and under-covered the wet regime at 0.744, with the same width in both by construction. CQR produced 0.886 coverage at width 0.01094\,$\mathrm{m^{3}\,m^{-3}}$ in the dry regime and 0.864 coverage at width 0.02371\,$\mathrm{m^{3}\,m^{-3}}$ in the wet regime. The seasonal breakdown showed the same pattern, with split conformal coverage falling to 0.730 in winter against 0.832 for CQR.

\begin{table}[htbp]
\centering
\caption{Conditional coverage and width by regime and season at the 24\,h horizon}
\label{tab:5}
\small
\resizebox{\linewidth}{!}{%
\begin{tabular}{lrrrrr}
\toprule
Slice & n & Split coverage & Split width & CQR coverage & CQR width \\
\midrule
All & 3\,713 & 0.826 & 0.01939 & 0.875 & 0.01725 \\
Wet & 1\,834 & 0.744 & 0.01939 & 0.864 & 0.02371 \\
Dry & 1\,879 & 0.906 & 0.01939 & 0.886 & 0.01094 \\
December--February & 1\,543 & 0.730 & 0.01939 & 0.832 & 0.02030 \\
March--May & 2\,170 & 0.894 & 0.01939 & 0.905 & 0.01508 \\
\bottomrule
\end{tabular}
}
\end{table}

Figure 5 shows the forecast and its interval over the final stretch of the test period, and Figure 6 shows calibration and sharpness directly.

\begin{figure}[htbp]
\centering
\includegraphics[width=0.94\linewidth]{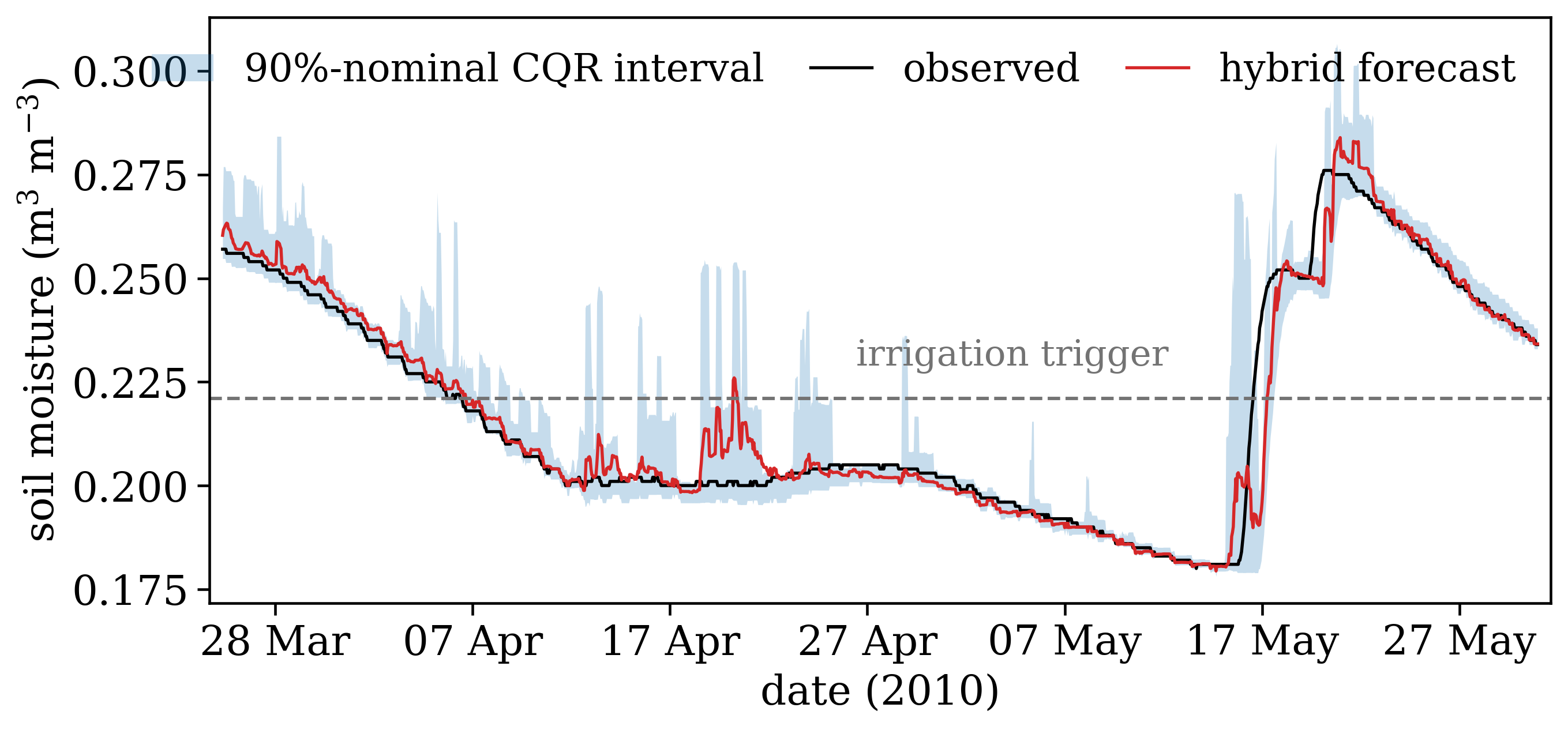}
\caption{Observed and forecast soil moisture with the 90\,\% conformalised interval over the last 1\,600\,h of the test partition. The spring drydown crosses the management trigger in early April; the band widens markedly around the rainfall event of mid-May, when the state is least predictable.}
\label{fig:5}
\end{figure}

\begin{figure}[htbp]
\centering
\includegraphics[width=1.00\linewidth]{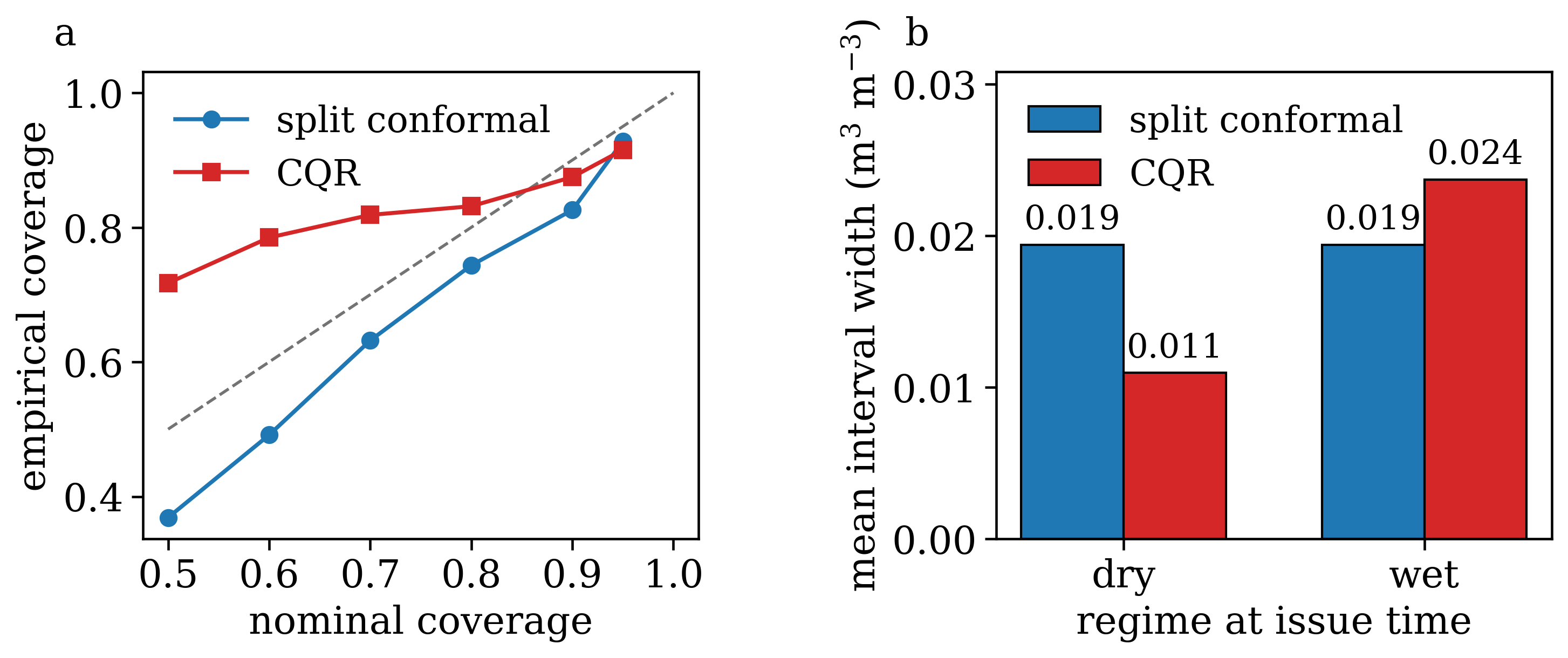}
\caption{Interval quality at the 24\,h horizon. (a) Empirical against nominal coverage; the dashed line marks perfect calibration. (b) Mean interval width by regime at issue time; the adaptive method narrows in dry conditions and widens in wet ones, whereas the constant-width method cannot.}
\label{fig:6}
\end{figure}

Across the horizon sweep the mean interval width grew by a factor of 40.5 from one hour to one week, while coverage remained between 0.855 and 0.940 (Figure 8a). The advantage of CQR was itself horizon-dependent: at one and three hours the constant-width method was equally well calibrated, while from 24\,h onward CQR was both better covered and sharper, by 19\,\% in width at 72\,h.

\subsection{Decision performance}

At the 24\,h horizon the adverse event, a management-threshold crossing, occurred in 25.9\,\% of test hours. The reactive rule missed 24 crossings, the point-forecast rule missed nine and issued 20 warnings a full 24\,h before the reactive rule responded, and the risk-aware rule missed none. The improvement was not free: precision fell from 0.975 to 0.950 and the notional water proxy rose 3.7\,\% above the reactive rule (Table 6). The comparison is therefore a risk-versus-water trade-off, not a water saving.

\begin{table}[htbp]
\centering
\caption{Decision-rule comparison at the 24\,h horizon}
\label{tab:6}
\small
\resizebox{\linewidth}{!}{%
\begin{tabular}{lrrrrrr}
\toprule
Rule & Alarms & Precision & Recall & F1 & Notional water (mm) & Missed crossings \\
\midrule
Reactive (no forecast) & 961 & 0.975 & 0.975 & 0.975 & 24\,397 & 24 \\
Forecast (point) & 969 & 0.982 & 0.991 & 0.987 & 24\,493 & 9 \\
\textbf{Risk-aware (interval lower bound)} & 1\,012 & 0.950 & \textbf{1.000} & 0.974 & 25\,309 & \textbf{0} \\
\bottomrule
\end{tabular}
}
\par\vspace{2pt}
\begin{minipage}{\linewidth}\footnotesize \emph{Note.} The water column is the notional refill-to-field-capacity proxy defined in Section 2.7, accumulated over every alarm hour of the 3\,713-hour test partition with no actuation and no state update. These totals are not field irrigation depths and are interpretable only as ratios between rules.\end{minipage}
\end{table}

Within the decision-relevant regime, comprising 390\,h and 252 crossings, the share detected in advance rose from 0.905 for the reactive rule to 0.984 for the point forecast and 1.000 for the risk-aware rule (Figure 7).

\begin{figure}[htbp]
\centering
\includegraphics[width=1.00\linewidth]{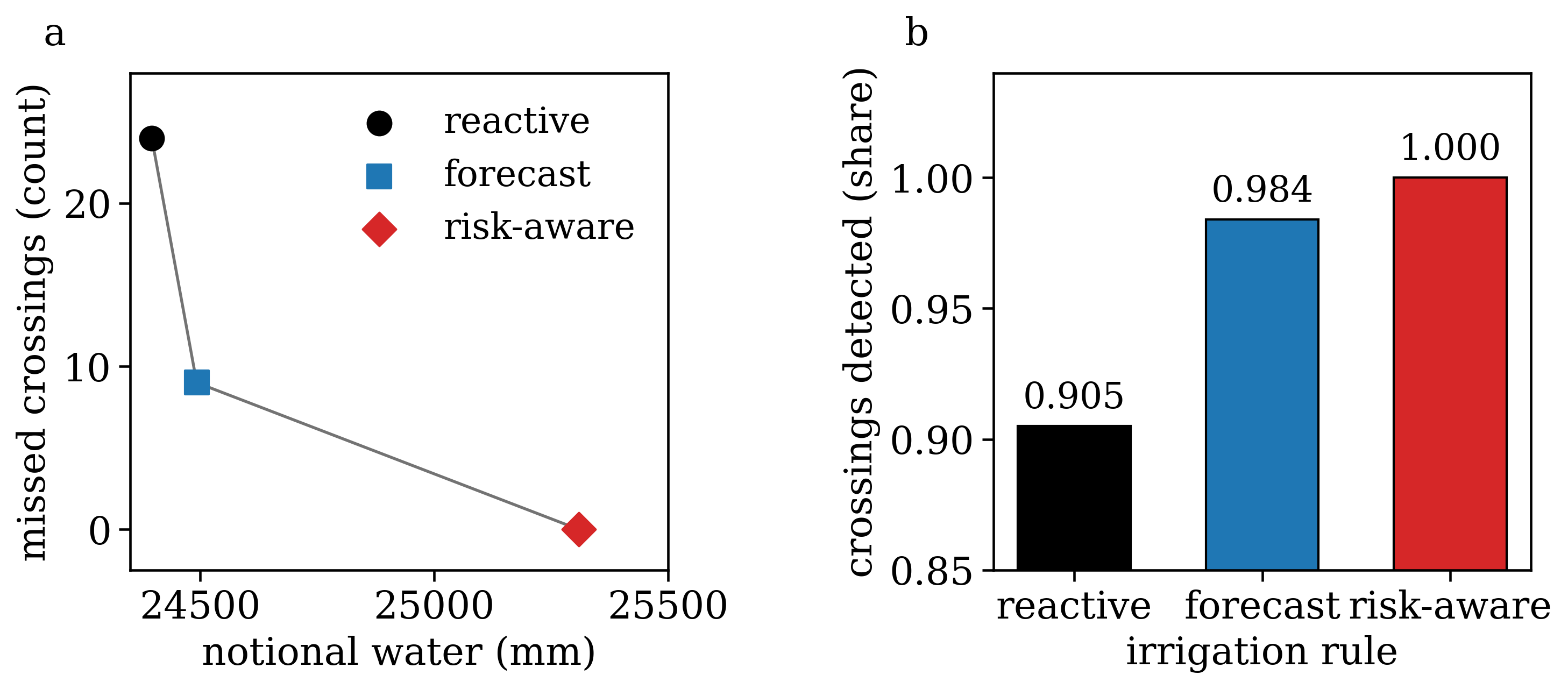}
\caption{Decision performance at the 24\,h horizon. (a) Notional water proxy against missed management-threshold crossings for the three rules. (b) Share of crossings detected in advance within the decision-relevant regime, where irrigation timing is genuinely difficult.}
\label{fig:7}
\end{figure}

Across lead times the three rules diverged (Figure 8b). The reactive rule degraded steadily from 0.996 at one hour to 0.616 at one week. The point-forecast rule degraded faster and \textbf{fell below the reactive rule beyond 72\,h}, reaching 0.710 against 0.760 at 72\,h and 0.581 against 0.616 at one week. The risk-aware rule retained a value of 1.000 at every horizon tested, at a notional water cost rising from 24\,785\,mm at one hour to 28\,995\,mm at one week.

\begin{figure}[htbp]
\centering
\includegraphics[width=1.00\linewidth]{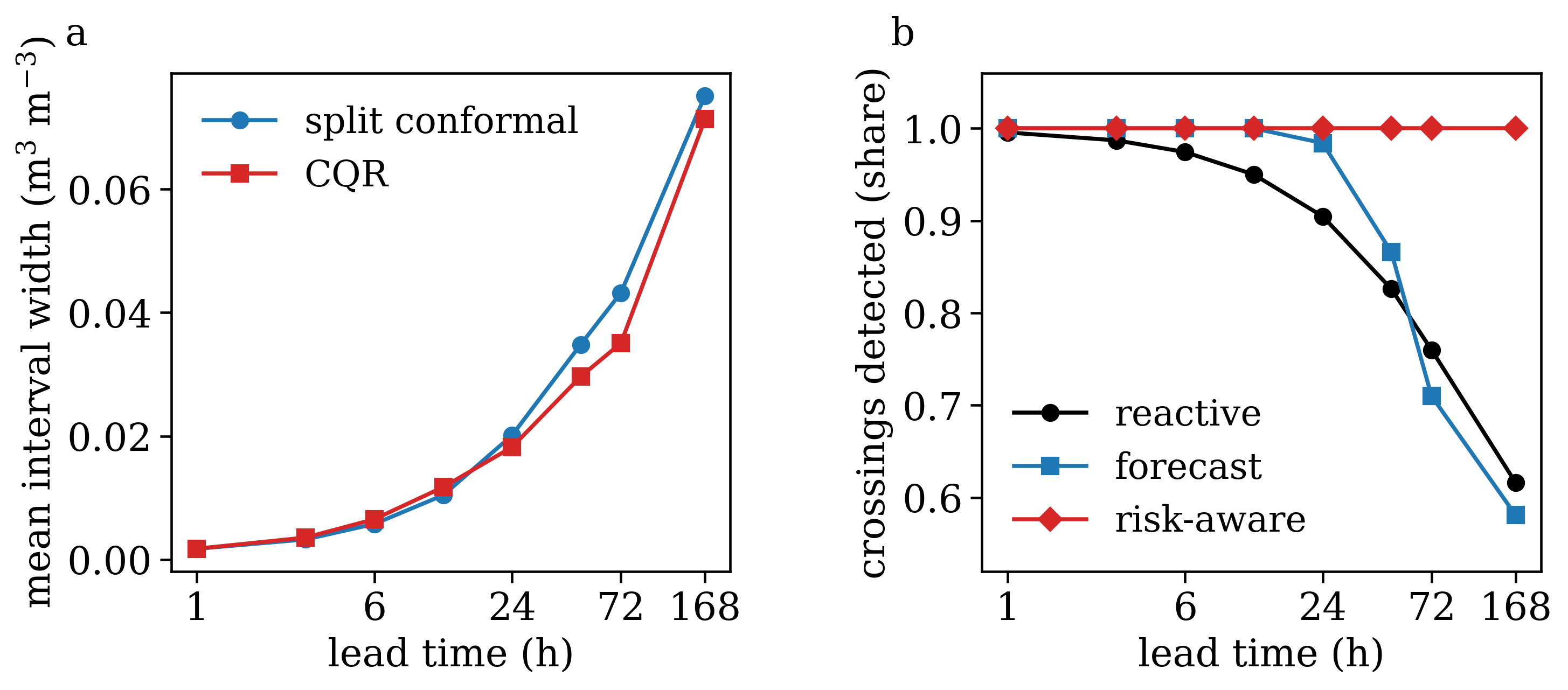}
\caption{Uncertainty and decisions against lead time. (a) Mean interval width; both methods widen as the forecast reaches further ahead, with the adaptive method consistently sharper beyond 24\,h. (b) Share of management-threshold crossings detected by each rule; the point forecast falls below the no-forecast rule beyond 72\,h whereas the interval-based rule does not.}
\label{fig:8}
\end{figure}

A selectivity diagnostic qualified this last result (Table 7). Because the rule fires whenever the interval lower bound falls below the trigger, and because the margin between trigger and lower bound was positive for every true crossing in the regime, perfect recall is partly structural. The informative quantity is the share of the regime on which the rule fires, together with precision. At 24\,h the rule declined to fire on 29\,\% of the regime and retained precision of 0.950. At 168\,h the interval half-width, 0.0357\,$\mathrm{m^{3}\,m^{-3}}$, exceeded the half-width of the decision regime itself, 0.02\,$\mathrm{m^{3}\,m^{-3}}$, the rule fired on 100\,\% of the regime, and precision fell to 0.704 with an alarm rate of 0.371 against an event base rate of 0.261.

\begin{table}[htbp]
\centering
\caption{Selectivity of the risk-aware rule across lead times}
\label{tab:7}
\small
\begin{tabular}{lrrrr}
\toprule
Lead time (h) & Interval half-width ($\mathrm{m^{3}\,m^{-3}}$) & Fires on regime (\%) & Precision & Alarm rate \\
\midrule
1 & 0.0009 & 60.0 & 0.979 & 0.263 \\
24 & 0.0091 & 71.0 & 0.950 & 0.273 \\
72 & 0.0175 & 92.7 & 0.888 & 0.292 \\
168 & 0.0357 & 100.0 & 0.704 & 0.371 \\
\bottomrule
\end{tabular}
\par\vspace{2pt}
\begin{minipage}{\linewidth}\footnotesize \emph{Note.} The adverse-event base rate is approximately 0.26 at all horizons.\end{minipage}
\end{table}

\subsection{Secondary and negative results}

Three further experiments sit outside the main line of the study and are reported here for completeness; none changes the results above.

\emph{Semi-supervised learning did not help.} Tri-training with three diverse learners and an unlabelled pool of 20\,951 feature rows drawn from two other CALABRIA stations reached +6.3\,\% skill at 24\,h, below the simple tuned Random Forest at +8.4\,\%, and per-learner error did not improve across rounds.

\emph{Feature reduction was nearly free but not beneficial.} Restricting the model to the eight highest-ranked features retained +3.2\,\% skill, within 0.0002\,$\mathrm{m^{3}\,m^{-3}}$ RMSE of the nineteen-feature baseline, so the additional predictors carry little independent information without being harmful.

\emph{Skill varied strongly with sensor depth.} At 30\,cm the tuned model achieved +8.4\,\% and at 90\,cm +20.6\,\%, but at 60\,cm it achieved \textbf{$-$41.1\,\%}: persistence RMSE at that depth was 0.00530\,$\mathrm{m^{3}\,m^{-3}}$ against 0.00748\,$\mathrm{m^{3}\,m^{-3}}$ for the model. The 60\,cm layer is slow and smooth enough that persistence is effectively unbeatable there, which is why skill is reported per depth rather than pooled across the profile (Section 4.5).

\section{Discussion}

\subsection{Why skill peaks at short lead time}

The inverted-U shape of the skill curve was not anticipated. The intuition that motivated the sweep was that persistence weakens with lead time and the hybrid should therefore gain. That is only half correct. Persistence does weaken, but the hybrid weakens in step with it beyond roughly one day, and the two converge.

The explanation lies in what the predictors contain. The model's advantage over persistence comes from resolving short-term dynamics that are \emph{already in motion} at issue time: a wetting front in transit through the profile, drainage in progress, an established drying trajectory. These processes leave signatures in the lags and rainfall accumulations and are therefore learnable, and they exert their influence over a few hours. Beyond about a day, the trajectory is increasingly governed by rainfall that has not yet fallen and is, by construction, absent from the feature set. No learner can recover information that is not present in its inputs.

This interpretation is supported independently by the perfect-forcing ablation at 24\,h. Supplying the observed future rainfall to the same architecture raised skill from +9.4\,\% to +33.4\,\%, which bounds how much of the residual error is attributable to the unknown forcing rather than to model capacity under this configuration. The skill-decay curve of Figure 3b is therefore consistent with rainfall uncertainty coming to dominate the error budget as lead time grows, and it indicates a practical ceiling of one to two days for the retrospective feature set used here. The ablation is strong evidence for that reading, but it is a single-site, single-configuration result and does not establish a universal causal limit on retrospective features.

\subsection{What the physical layer contributes}

The physical core was not a competitive standalone predictor even under its best calibration: +3.1\,\% over persistence at 24\,h in the deployable protocol adopted throughout (Section 3.2), against +9.4\,\% for the hybrid built on it. A sensitivity observation sharpens the point. Restricting the drainage coefficient to $k \le 0.4$ yielded that +3.1\,\%, whereas permitting $k \le 0.8$ selected $k = 0.60$, which fitted the training partition better but generalised worse, yielding +0.35\,\%; under the wider grid used in the horizon sweep the core stayed within $\pm$1.6\,\% of persistence at every lead time. What matters is that the \textbf{hybrid was unaffected}, returning RMSE 0.00925\,$\mathrm{m^{3}\,m^{-3}}$ in both cases, because the residual learner absorbed the difference. The accuracy of the end-to-end forecast is thus robust to a calibration choice that moves the standalone core by an order of magnitude in relative skill, which is a stronger statement about the decomposition than any standalone score would be. The practical conclusion is that the transparent core should be presented as a \emph{learnable decomposition} of the problem (a physically meaningful baseline whose residual is structured and therefore correctable) rather than as a predictor competing on accuracy. Its value is that every term retains an interpretation, which is what an auditable recommendation requires (G1, G2).

Two parameter findings deserve emphasis. First, the evapotranspiration weight $\beta$ calibrated to zero not only at 24\,h but at every horizon up to one week. It was plausible that evapotranspiration, negligible over a day, would begin to matter over multi-day windows; it did not. The useful physics therefore reduces to current state plus drainage above field capacity. Because the Hargreaves formulation used here is temperature-only, this finding may partly reflect the crudeness of that estimate rather than a genuine absence of evapotranspiration signal, and the two explanations cannot be separated with the available meteorological variables. Second, the drainage coefficient grew systematically with lead time while the field-capacity threshold drifted downward, both of which are physically sensible: more water drains as the forecast window lengthens, and the effective threshold at which drainage becomes active shifts as the accumulation period grows.

\subsection{Uncertainty: what is gained and where the claim stops}

The uncertainty layer behaved as a well-specified estimate should. Width scaled by a factor of 40.5 across the horizon range, so the system communicates without prompting that a one-week forecast is far less certain than a one-hour one. More importantly, the conditional audit showed that constant-width intervals have the \emph{wrong shape} for this problem: they over-cover dry conditions and under-cover wet ones, because the error is heteroscedastic and its heteroscedasticity is driven by rainfall. The adaptive method reshapes the interval to the state, widening precisely when the physics says the forecast is at risk. This closes the loop with Section 4.1: the same unknown-rainfall mechanism that limits point accuracy is what the interval must, and does, encode (G4).

Neither method reached nominal coverage on test, which is why the intervals are described throughout as 90\,\%-\emph{nominal} rather than as calibrated. The shortfall is consistent with temporal and seasonal distribution shift: conformal guarantees rest on exchangeability, the calibration partition falls in a different season from the test partition, and the observed loss of coverage is largest in winter, which is the direction that mechanism predicts. The present design cannot separate that mechanism from other departures from exchangeability, so the reading is offered as consistent with the evidence rather than demonstrated by it. It is reported rather than concealed, and it identifies rolling or adaptive conformal calibration as the natural next step.

The decision results require a similar qualification, which is worth stating precisely because the headline number invites over-reading. The risk-aware rule retained a perfect catch rate at every horizon, but this is \emph{partly structural}: for every true crossing inside the decision regime the margin between the trigger and the interval lower bound was positive by construction, rising from 0.015\,$\mathrm{m^{3}\,m^{-3}}$ at one hour to 0.041\,$\mathrm{m^{3}\,m^{-3}}$ at one week, so the rule fires by design. Table 7 shows what this costs. At 24\,h the result is substantive: the rule still discriminates, declining to fire on 29\,\% of the regime, and precision remains at 0.950 against 0.975 for the reactive rule, so the rise from 0.905 to 1.000 is bought for roughly 2.5 points of precision. By 72\,h the rule fires on 92.7\,\% of the regime and is becoming conservative by brute force. At one week the interval half-width exceeds the half-width of the decision regime itself, the rule fires everywhere within it, perfect recall becomes tautological, and precision collapses to 0.704. \textbf{The perfect-recall claim should therefore be made for lead times up to roughly one to two days and withdrawn at one week.} What survives at long lead is weaker but still meaningful: the interval-based rule fails \emph{safe}, whereas the point forecast fails \emph{silently}, dropping below even the no-forecast rule beyond 72\,h.

\subsection{Relation to previous findings}

The result that only the bagged tree ensemble beat persistence is consistent with the emphasis in Zheng et al. (2024) on short-horizon gains, and with the general observation that deep sequence models require either more data or richer forcing than a single station provides. The failure of Gaussian-process regression is a scale artefact: exact inference is cubic in sample size, so the model was fitted to a subsample. It should not be read as evidence against the method class.

The finding that a hybrid outperforms its components agrees in direction with Li et al. (2024), who reported that physics-informed deep learning beat pure deep learning most clearly at long lead times. The present study differs in that its physical component is deliberately minimal, four parameters rather than a land-surface model, and in that the improvement here was concentrated at \emph{short} lead time. The two results are compatible: Li et al. (2024) supplied their networks with physically simulated forcing that carries information about the future, whereas the deployable configuration here has no future information at all, a difference consistent with the decay seen in Figure 3b.

Izquierdo-Sanz and Molt\'o (2026) worked in the closest setting, hybrid physical--machine-learning modelling of Mediterranean irrigated soil, but addressed estimation of current moisture from satellite imagery rather than forecasting, and attached neither uncertainty nor a decision rule. Zheng et al. (2026) came closest on the decision side, coupling Monte-Carlo dropout uncertainty to an irrigation index and reporting water savings. The distinction is the nature of the uncertainty: Monte-Carlo dropout measures disagreement within an assumed model class and provides no finite-sample coverage guarantee, whereas the conformal construction used here provides a distribution-free guarantee under exchangeability and, as Section 4.3 shows, permits that guarantee to be \emph{tested} and found wanting under seasonal shift. Tandon et al. (2022) similarly attached model-based uncertainty without carrying it to a decision.

On interpretability, the drivers recovered here (current state, short lags, antecedent wetness, a small temperature and seasonal signal) are consistent with the physically meaningful drivers reported by Huang et al. (2023) for soil-moisture drought, though their site-level analysis emphasised soil temperature and atmospheric aridity, variables not available at this station. The divergence between permutation importance and Shapley ranking of the rainfall features is expected under collinearity: permutation importance measures marginal contribution when a single column is destroyed, which disadvantages correlated predictors, while Shapley values distribute credit among them. Both identified rainfall accumulation as the leading non-soil driver, which is the agronomically relevant conclusion. It should be read for what it is: an explanation of the predictive component. The recommendation itself is not explained by feature attribution but is interpretable by construction, because the action follows a stated comparison between an interval bound and a documented management threshold. That separation, attribution for the forecast and an explicit rule for the decision, is what closes G3 here, and it is a different claim from the one usually made when Shapley analysis is offered as an explanation of a decision.

\subsection{Limitations}

Five limitations bound the conclusions. First, evaluation was \textbf{open-loop}: no irrigation actuation exists in the data, so the water figure is a notional refill proxy and a missed event is a failure to give early warning of a management-threshold crossing, not an observed crop outcome. The adverse event is defined on soil moisture alone; no physiological measurement of crop stress is available at this station. Second, the chronological split placed the test partition between December and May, the \textbf{low-demand half-year} for irrigation; summer, when the decision matters most, is under-exercised, and this is the most consequential gap for operational credibility. Third, \textbf{interval coverage fell below nominal} under seasonal shift. Fourth, evapotranspiration was estimated from \textbf{temperature alone}, which confounds the interpretation of $\beta$ = 0. Fifth, the main comparison used a \textbf{single station, a single depth and a single tuned hyperparameter configuration} reused across horizons; a pooled cross-station model indicated that absolute soil moisture depends strongly on soil porosity, so spatial transfer requires reformulating the target rather than retraining, and per-horizon tuning could shift the skill curve, most plausibly at its extremes.

The depth result is a limitation of scope but also a finding in its own right: at 60\,cm the layer is sufficiently slow and smooth that persistence is effectively unbeatable, and the model degraded skill by 41.1\,\%. Reporting a single headline skill for a profile would therefore be misleading, and skill should be reported per depth.

\subsection{Implications and recommendations}

Three implications follow for practice. First, \textbf{transparency did not cost measurable accuracy in this configuration}: the tuned hybrid at +9.4\,\% and the tuned black-box Random Forest at +8.4\,\% are within 0.0001\,$\mathrm{m^{3}\,m^{-3}}$ RMSE of each other, so the contribution is not that the transparent model wins by a wide margin but that comparable accuracy is obtainable inside a physical--residual decomposition that confines opacity to a bounded correction term and to which conformal intervals and an explicit decision rule attach naturally. On this evidence auditable architectures are a reasonable default rather than a concession, though the comparison is a single-site one. Second, the useful range of a point forecast is short, one to two days on this configuration, and claims of multi-day soil-moisture forecasting skill should be checked against persistence at the same horizon before being acted upon. Third, and most consequentially for decision support, \textbf{uncertainty is what extends usable range}: beyond the point forecast's useful horizon the interval-based rule remains safe where the point forecast becomes actively misleading, and the price of that safety is explicit and monotone in water.

Four extensions follow directly. \textbf{Rainfall-forecast forcing} from numerical weather prediction or reanalysis is the most direct route to extending range indicated by the present results, the perfect-forcing ablation being the evidence, and should sharpen intervals and decisions simultaneously. \textbf{Rolling or adaptive conformal calibration} should restore nominal coverage under seasonal shift. \textbf{Summer-inclusive validation} is required before any operational claim. Finally, \textbf{multi-station transfer} should be pursued through a change-of-variable formulation rather than by pooling absolute moisture, given the porosity dependence observed here.

\section{Conclusions}

This study coupled a four-parameter water-balance core to a machine-learning residual corrector, attached distribution-free conformal intervals, and converted those intervals into an irrigation decision, evaluating the whole chain on three years of hourly Mediterranean in-situ measurements under a strict chronological protocol.

Four conclusions follow. First, the transparent hybrid answered RQ1 affirmatively: it reached +9.4\,\% skill against persistence at 24\,h, roughly double the best of nine baselines, of which only the Random Forest beat persistence at all; the margin over that tuned black box was itself small (+9.4\,\% against +8.4\,\%), so the substance of the result is that this accuracy was obtained inside an auditable physical--residual decomposition rather than that transparency bought accuracy. Second, RQ2 was satisfied structurally rather than post hoc: the physical layer is interpretable by construction and the irrigation action follows an explicit threshold rule, while permutation and Shapley analysis account for the predictive component and show it resting on agronomically meaningful drivers. Third, RQ3 was answered as a trade-off with a boundary: within a one- to two-day range the risk-aware rule detected every management-threshold crossing in the decision-relevant regime while remaining selective and precise, at a cost of 2.5 points of precision and 3.7\,\% in the notional water proxy, so it buys earlier detection rather than saving water; beyond that range it remained safe but ceased to discriminate, and the perfect-recall claim must be bounded accordingly. Fourth, and most broadly, forecast skill peaked at +27.4\,\% at three hours and decayed to +1.2\,\% at one week, a decay consistent with unknown future rainfall coming to dominate the error budget and indicating a ceiling for the retrospective feature set used here.

The central implication is that in this problem the quality of the uncertainty estimate, rather than point accuracy, governs how far ahead a forecast can responsibly guide an irrigation decision. On the present evidence, extending that range points to information about future rainfall rather than to a better learner.

\section*{Acknowledgements}

The author thanks Somyajit Chakraborty for research supervision. Soil-moisture, temperature and precipitation data were obtained from the International Soil Moisture Network; the author thanks the network and the data-providing agency for making the CALABRIA records publicly available. Static soil properties were obtained from the Harmonized World Soil Database.

\section*{Data and code availability}

The measurement data are publicly available from the International Soil Moisture Network. The full analysis code, comprising the water-balance core, the residual learner, the conformal calibration layer and the scripts that regenerate every figure and table, is openly available at \url{https://github.com/ascariolo/hybrid-model-irrigation-dss} under the MIT licence. That repository also contains the executed notebooks and the tabulated results underlying each figure and table, together with instructions for retrieving the CALABRIA station records from the International Soil Moisture Network.

\section*{References}
\begin{sloppypar}
\noindent\hangindent=1.5em\hangafter=1 Ahmed, A. A. M., Deo, R. C., Raj, N., Ghahramani, A., Feng, Q., Yin, Z., \& Yang, L. (2021). Deep learning forecasts of soil moisture: Convolutional neural network and gated recurrent unit models coupled with satellite-derived MODIS, observations and synoptic-scale climate index data. \emph{Remote Sensing, 13}(4), 554. \url{https://doi.org/10.3390/rs13040554}\par\vspace{4pt}
\noindent\hangindent=1.5em\hangafter=1 Ajith, S., Vijayakumar, S., \& Elakkiya, N. (2025). Yield prediction, pest and disease diagnosis, soil fertility mapping, precision irrigation scheduling, and food quality assessment using machine learning and deep learning algorithms. \emph{Discover Food, 5}(1), Article 67. \url{https://doi.org/10.1007/s44187-025-00338-1}\par\vspace{4pt}
\noindent\hangindent=1.5em\hangafter=1 Bakhshian, S., Zarepakzad, N., Nevermann, H., Hohenegger, C., Or, D., \& Shokri, N. (2025). Field-scale soil moisture dynamics predicted by deep learning. \emph{Advances in Water Resources, 201}, Article 104976. \url{https://doi.org/10.1016/j.advwatres.2025.104976}\par\vspace{4pt}
\noindent\hangindent=1.5em\hangafter=1 El Rhadiouini, C., Jin, S., Yeboah, E., Sarfo, I., \& Okrah, A. (2026). A hybrid physics-guided machine learning for soil moisture monitoring and drought assessment from CYGNSS in complex terrain. \emph{IEEE Transactions on Geoscience and Remote Sensing, 64}, 1--19. \url{https://doi.org/10.1109/TGRS.2026.3677267}\par\vspace{4pt}
\noindent\hangindent=1.5em\hangafter=1 Food and Agriculture Organization of the United Nations. (n.d.). \emph{AQUASTAT: Water withdrawal by sector.} \url{https://www.fao.org/aquastat/}\par\vspace{4pt}
\noindent\hangindent=1.5em\hangafter=1 Goap, A., Sharma, D., Shukla, A. K., \& Krishna, C. R. (2018). An IoT based smart irrigation management system using machine learning and open source technologies. \emph{Computers and Electronics in Agriculture, 155}, 41--49. \url{https://doi.org/10.1016/j.compag.2018.09.040}\par\vspace{4pt}
\noindent\hangindent=1.5em\hangafter=1 Huang, F., Zhang, Yongkun, Zhang, Ye, Nourani, V., Li, Q., Li, L., \& Shangguan, W. (2023). Towards interpreting machine learning models for predicting soil moisture droughts. \emph{Environmental Research Letters, 18}(7), Article 074002. \url{https://doi.org/10.1088/1748-9326/acdbe0}\par\vspace{4pt}
\noindent\hangindent=1.5em\hangafter=1 Islam, M. B., Guerrieri, A., Gravina, R., Delaney, D. T., \& Fortino, G. (2025). From traditional machine learning to fine-tuning large language models: A review for sensors-based soil moisture forecasting. \emph{Sensors, 25}(22), 6903. \url{https://doi.org/10.3390/s25226903}\par\vspace{4pt}
\noindent\hangindent=1.5em\hangafter=1 Izquierdo-Sanz, H., \& Molt\'o, E. (2026). Hybrid physical--machine learning soil moisture modeling at orchard scale in irrigated citrus orchards using Sentinel 1 and 2 and agroclimatic data. \emph{Agronomy, 16}(5), 541. \url{https://doi.org/10.3390/agronomy16050541}\par\vspace{4pt}
\noindent\hangindent=1.5em\hangafter=1 Khaliq, A., Khan, A., Jan, S., Umair, M., Gulshair, A., Ali, A., \& Ali Shah, U. (2025). AI-driven smart agriculture: An integrated approach for soil analysis, irrigation, and crop-fertilizer recommendations. \emph{IEEE Access, 13}, 141124--141138. \url{https://doi.org/10.1109/ACCESS.2025.3594162}\par\vspace{4pt}
\noindent\hangindent=1.5em\hangafter=1 Koohikeradeh, E., Gumiere, S. J., \& Bonakdari, H. (2025). NDMI-derived field-scale soil moisture prediction using ERA5 and LSTM for precision agriculture. \emph{Sustainability, 17}(6), 2399. \url{https://doi.org/10.3390/su17062399}\par\vspace{4pt}
\noindent\hangindent=1.5em\hangafter=1 Li, L., Dai, Y. J., Wei, Z. W., Shangguan, W., Wei, N., Zhang, Y. G., Li, Q. L., \& Li, X.-X. (2024). Enhancing deep learning soil moisture forecasting models by integrating physics-based models. \emph{Advances in Atmospheric Sciences, 41}(7), 1326--1341. \url{https://doi.org/10.1007/s00376-023-3181-8}\par\vspace{4pt}
\noindent\hangindent=1.5em\hangafter=1 Liakos, K. G., Busato, P., Moshou, D., Pearson, S., \& Bochtis, D. (2018). Machine learning in agriculture: A review. \emph{Sensors, 18}(8), 2674. \url{https://doi.org/10.3390/s18082674}\par\vspace{4pt}
\noindent\hangindent=1.5em\hangafter=1 Mallik, S., Chakraborty, A., Podder, K., Talukdar, S., Rahman, A., \& Mishra, U. (2025). Enhancing soil moisture prediction with explainable AI: Integrating IoT and multi-sensor remote sensing data through soft computing. \emph{Applied Soft Computing, 180}, Article 113406. \url{https://doi.org/10.1016/j.asoc.2025.113406}\par\vspace{4pt}
\noindent\hangindent=1.5em\hangafter=1 Mohan, R. N. V. J., Rayanoothala, P. S., \& Sree, R. P. (2025). Next-gen agriculture: Integrating AI and XAI for precision crop yield predictions. \emph{Frontiers in Plant Science, 15}, Article 1451607. \url{https://doi.org/10.3389/fpls.2024.1451607}\par\vspace{4pt}
\noindent\hangindent=1.5em\hangafter=1 Nguyen, T. T., Ngo, H. H., Guo, W., Chang, S. W., Nguyen, D. D., Nguyen, C. T., Zhang, J., Liang, S., Bui, X. T., \& Hoang, N. B. (2022). A low-cost approach for soil moisture prediction using multi-sensor data and machine learning algorithm. \emph{Science of the Total Environment, 833}, Article 155066. \url{https://doi.org/10.1016/j.scitotenv.2022.155066}\par\vspace{4pt}
\noindent\hangindent=1.5em\hangafter=1 Settu, P., \& Ramaiah, M. (2025). A data driven comparison of hybrid machine learning techniques for soil moisture modeling using remote sensing imagery. \emph{Scientific Reports, 15}, Article 43170. \url{https://doi.org/10.1038/s41598-025-27225-0}\par\vspace{4pt}
\noindent\hangindent=1.5em\hangafter=1 Shahriari, B., Swersky, K., Wang, Z., Adams, R. P., \& de Freitas, N. (2016). Taking the human out of the loop: A review of Bayesian optimization. \emph{Proceedings of the IEEE, 104}(1), 148--175. \url{https://doi.org/10.1109/JPROC.2015.2494218}\par\vspace{4pt}
\noindent\hangindent=1.5em\hangafter=1 Taheri, M., Bigdeli, B., Imanian, H., \& Mohammadian, A. (2025). An overview of machine-learning methods for soil moisture estimation. \emph{Water, 17}(11), 1638. \url{https://doi.org/10.3390/w17111638}\par\vspace{4pt}
\noindent\hangindent=1.5em\hangafter=1 Tandon, K., Sen, S., Kasiviswanathan, K. S., Soundharajan, B. S., Tummuru, N. R., \& Das, A. (2022). Integration of machine learning and particle filter approaches for forecasting soil moisture. \emph{Stochastic Environmental Research and Risk Assessment, 36}(12), 4235--4253. \url{https://doi.org/10.1007/s00477-022-02258-3}\par\vspace{4pt}
\noindent\hangindent=1.5em\hangafter=1 Wang, Y., Shi, L., Hu, Y., Hu, X., Song, W., \& Wang, L. (2024). A comprehensive study of deep learning for soil moisture prediction. \emph{Hydrology and Earth System Sciences, 28}, 917--943. \url{https://doi.org/10.5194/hess-28-917-2024}\par\vspace{4pt}
\noindent\hangindent=1.5em\hangafter=1 Xu, L., Yu, H., Chen, Z., Du, W., Chen, N., \& Huang, M. (2023). Hybrid deep learning and S2S model for improved sub-seasonal surface and root-zone soil moisture forecasting. \emph{Remote Sensing, 15}(13), 3410. \url{https://doi.org/10.3390/rs15133410}\par\vspace{4pt}
\noindent\hangindent=1.5em\hangafter=1 Zheng, W., Zheng, K., Gao, L., Zhangzhong, L., Lan, R., Xu, L., \& Yu, J. (2024). GRU--Transformer: A novel hybrid model for predicting soil moisture content in root zones. \emph{Agronomy, 14}(3), 432. \url{https://doi.org/10.3390/agronomy14030432}\par\vspace{4pt}
\noindent\hangindent=1.5em\hangafter=1 Zheng, D., Yang, C., Zheng, G., Belgibaev, B., Mansurova, M., Jomartova, S., \& Zhao, B. (2026). Informer--UNet: A hybrid deep learning framework for multi-point soil moisture prediction and precision irrigation in winter wheat. \emph{Agriculture, 16}(6), 648. \url{https://doi.org/10.3390/agriculture16060648}\par\vspace{4pt}
\end{sloppypar}

\end{document}